\documentclass{article}

\PassOptionsToPackage{numbers,compress}{natbib}
\usepackage[preprint]{neurips_2026}
\usepackage{graphicx}
\usepackage{amsmath}
\usepackage{enumitem}
\usepackage{subcaption}
\usepackage{tabularx}
\usepackage{listings}

\usepackage[utf8]{inputenc} 
\usepackage[T1]{fontenc}    
\usepackage{hyperref}       
\usepackage{url}            
\usepackage{booktabs}       
\usepackage{amsfonts}       
\usepackage{nicefrac}       
\usepackage{microtype}      
\usepackage{xcolor}         

\title{Why Better Models Can Create Riskier Systems: Evidence from LLM Agents in Financial Markets}

\author{
  Jillian Ross \\
  Department of Electrical Engineering and Computer Science\\
  Massachusetts Institute of Technology \\
  \texttt{jillianr@mit.edu} \\
  \AND
  Eric So \\
  Sloan School of Management \\
  Massachusetts Institute of Technology \\
  \texttt{eso@mit.edu} \\
  \AND
  Zoe De Simone \\
  Department of Electrical Engineering and Computer Science \\
  Massachusetts Institute of Technology \\
  \texttt{zoed@mit.edu} \\
  \AND
  Charles Pozniak \\
  Belfer Center for Science and International Affairs \\
  Harvard Kennedy School \\
  \texttt{cpozniak@hks.harvard.edu} \\
  \AND
  Andrew W. Lo \\
  Sloan School of Management \\
  Massachusetts Institute of Technology \\
  \texttt{alo-admin@mit.edu} \\
}

\begin{document}

\maketitle

\begin{abstract}
Large language models (LLMs) are being deployed at scale in consequential real-world systems, from financial markets to content moderation to hiring. We show that improving individual model capability can degrade rather than improve system-level outcomes. We hypothesize that shared training and architectures can lead more capable LLMs to behave more similarly, creating correlated actions that do not diversify away. We develop a general framework showing how this correlation creates a non-diversifiable risk floor and test its predictions in financial markets using an agent-based simulation with LLM traders of varying general-purpose capability. We find that: (1) frontier LLMs exhibit significantly correlated behavior that increases with capability; (2) when their shared reasoning is accurate, increasing agent participation reduces market-level risk; and (3) when agents share a common misinformation environment, the same correlated behavior becomes a liability. Together, these results identify a capability paradox: improving individual models does not necessarily produce better system-level outcomes. Whether the same dynamics arise in other domains is an open empirical question.
\end{abstract}

\section{Introduction}

The deployment of large language models (LLMs) in high-stakes systems, from financial markets to content moderation to corporate hiring, raises a deceptively simple question: can improving individual model capability degrade rather than improve system-level outcomes?  When LLMs are embedded in multi-agent dynamic systems, the relationship between model quality and system-level behavior is not well understood. We argue that agent quality and collective behavior can decouple in ways that no individual evaluation can detect. Models that perform well in isolation may, when deployed at scale alongside similar models, produce collective behavior that destabilizes the very system they were designed to improve. 

We decompose each agent's action into a \emph{corrective} component---pressure that moves the system toward ground truth---and a \emph{non-corrective} component---the residual behavior left after accounting for that pressure. This decomposition is functional. Non-corrective actions can reflect well-reasoned responses to narrative framing, learned heuristics, portfolio constraints, or priors that are informative on average but misaligned in a given context. Their system-level effect depends on correlation across agents. Independent non-corrective actions cancel in the aggregate; correlated ones persist as the population grows.

Consider a financial market populated by automated trading agents. Corrective actions push prices toward fundamental value. Correlated non-corrective actions can produce persistent mispricing, excess volatility, or shock amplification. We use \emph{systemic risk} to refer to widespread instability generated by correlated behavior that impairs normal system functioning \citep{european_central_bank_systemic_2009}.

We elect to empirically study this question in the context of financial markets, which offer a uniquely tractable testbed: a ground truth exists (the fundamental asset value), outcomes are continuous and measurable (price efficiency and volatility), and the consequences of poor collective behavior are well-characterized. Using agent-based market simulations, we replace algorithmic traders with LLM agents of varying capability and measure the resulting change in market-level properties. This setting allows us to address three research questions:

\begin{enumerate}
\item[\textbf{RQ1.}] Do frontier LLMs exhibit correlated non-corrective behavior?
\item[\textbf{RQ2.}] Does increasing LLM market participation produce systemic risk?
\item[\textbf{RQ3.}] Does a shared information environment amplify systemic risk?
\end{enumerate}

We find that frontier LLMs exhibit significantly correlated non-corrective behavior, and this correlation grows with model capability (RQ1). We also find that increasing LLM participation monotonically improves convergence toward the ground truth across all model families tested. Cross-model differences in convergence quality are well explained by how aggressively each model acts on perceived deviation from the ground truth (RQ2). However, when LLM agents operate within a shared information environment the stabilizing benefit disappears entirely: coordinated reliance on misaligned information can produce market dynamics that are strictly less stable than those generated by uninformed noise traders alone (RQ3). These results suggest that the correlation structure of model actions is a central determinant of systemic risk from LLM deployment. Because that correlation grows with capability, more capable models can pose greater systemic risk than less capable ones.

\section{Systemic Risk in LLM Agents}
\label{sec:theory}

Consider a system with a ground truth state $F \in \mathbb{R}$ and an observable state $S_t \in \mathbb{R}$ evolving over discrete time $t = 0,1,\dots,T$. A population of agents takes actions that collectively update $S_t$, and system quality depends on how closely $S_t$ tracks $F$ over time. We develop the framework in this general form and then test its predictions in financial markets, where $F$ is fundamental asset value and $S_t$ is market price.
 
\subsection{Decomposing agent actions}
 
The system is populated by $N$ agents indexed $i \in \{1, \dots, N\}$, each observing $S_t$ and taking an action $a_t^{(i)} \in \mathbb{R}$. The system state evolves as:
 
\begin{equation}
    S_{t} = S_{t-1} + \frac{1}{N}\sum_{i=1}^N a_t^{(i)}
    \label{eq:state_evolution}
\end{equation} 
 
Each agent's action is a scalar encoding the direction and magnitude of pressure it applies to $S_t$. We split each action into a \emph{corrective} component tied to the contemporaneous gap and a \emph{non-corrective} residual component:
 
\begin{equation}
    a^{(i)}_t = \underbrace{\lambda_i\, G_{t-1}}_{\text{corrective}} + \underbrace{r^{(i)}_t}_{\text{non-corrective}},
    \qquad
    \lambda_i := \frac{\mathrm{Cov}(a^{(i)}_t, G_{t-1})}{\mathrm{Var}(G_{t-1})}.
    \label{eq:decomp}
\end{equation}
 
$\lambda_i$ is the population regression coefficient of $a^{(i)}_t$ on the gap, and $r^{(i)}_t$ is by construction mean-zero and uncorrelated with $G_{t-1}$. Let $\bar\lambda := N^{-1}\sum_i \lambda_i$. This decomposition serves as a simple statistical framework for quantifying how similarly agents respond to shared information, as a statistical description of behavior. In the market setting developed later, the systematic policy can admit richer feature expansions; the decomposition here is the gap-based baseline.

For an LLM trader, a fundamental signal of \$120 with a current price of \$100 induces a corrective buy component tied to the \$20 gap. The residual captures the remaining response to features such as short-term trends, portfolio constraints, or market narratives.
 
\subsection{Correlated action creates a non-diversifiable floor}

\begin{figure}[t]
    \centering
    \vspace{-1em}
    \includegraphics[width=0.8\linewidth]{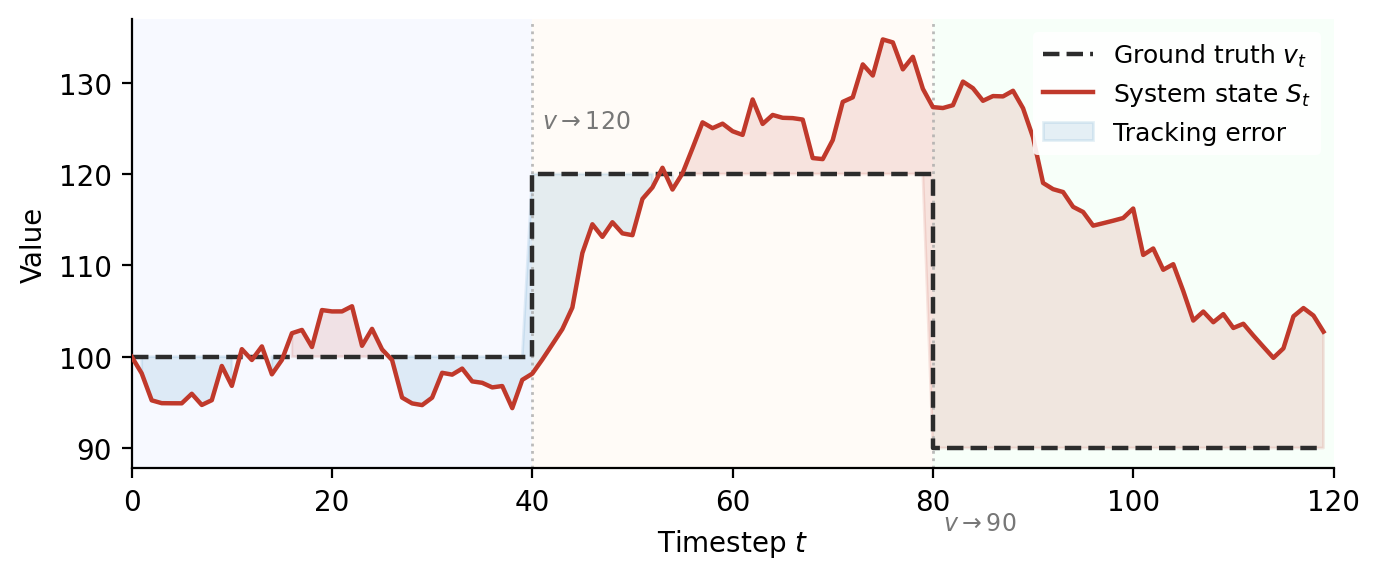}
    \caption{System state $S_t$ (red) tracking a ground truth $F$ (dashed) that shifts from 100 to 120 at $t=40$ and to 90 at $t=80$. Shaded regions show tracking error. Correlated non-corrective behavior across agents prevents $S_t$ from converging to $F$, producing a persistent floor on aggregate error.}
    \label{fig:simulation_fundamental}
\end{figure}
 
The key question is whether non-corrective behavior is independent or correlated across agents. If $r_t^{(i)}$ and $r_t^{(j)}$ of agents $i$ and $j$ are uncorrelated, their contributions cancel in the aggregate and system performance improves with $N$. If they are positively correlated, a component of each agent's action moves in the same direction simultaneously and cannot be diversified away, no matter how large the population. This pressure becomes systematically destabilizing when it prevents the state from returning to ground truth or amplifies deviations after shocks.

Formally, let $\bar\rho := \frac{1}{N(N-1)}\sum_{i \neq j}\mathrm{Corr}(r^{(i)}_t, r^{(j)}_t)$ denote the average pairwise correlation of non-corrective components. When $\bar\rho > 0$, the variance of the average non-corrective components decomposes as
\begin{equation}
    \mathrm{Var}(\bar r_t) = \underbrace{\frac{(1-\bar\rho)\sigma_r^2}{N}}_{\text{diversifiable}} + \underbrace{\bar\rho\,\sigma_r^2}_{\text{floor}},
\end{equation}
where the first term vanishes with population size and the second does not. The floor $\bar\rho\,\sigma_r^2$ is invariant in $N$: no amount of population scaling can eliminate it. 

\subsection{Estimating correlated action}
\label{subsec:identification}
 
We estimate correlated action as follows. For each agent $i$, we fit a separate OLS regression on a training set of scenarios:

\begin{equation}
    \hat{a}^{(i)} = \beta_0^{(i)} + \boldsymbol{\beta}^{(i)\top} \mathbf{x}
    \label{eq:ols}
\end{equation}

\noindent where $\mathbf{x}$ collects the observable state features. We then compute non-corrective component estimates $\hat{r}^{(i)} = a^{(i)} - \hat{a}^{(i)}$ on a disjoint test set --- the portion of each model's action not explained by its own fitted corrective component.

Correlated action between a pair of models $(i, j)$ is then estimated as the Pearson correlation of their non-corrective component across scenarios:

\begin{equation}
    \hat{\rho}_{ij}^{(r)} = \mathrm{Corr}\!\left(\hat{r}^{(i,r)},\,
    \hat{r}^{(j,r)}\right)
    \label{eq:rho_est}
\end{equation}

A high value of $\hat{\rho}_{ij}^{(r)}$ means models $i$ and $j$ exhibit correlated non-corrective components across scenarios---a signature of common action that population scaling cannot eliminate. We study how $\hat{\rho}_{ij}^{(r)}$ varies with model capability, and whether same-provider pairs exhibit higher correlation than cross-provider pairs. The formal decomposition of the non-corrective component into architecture-driven and portfolio-driven components, and how mixing across model families operates on each, is in Appendix~\ref{app:noise_decomposition}. We present a general framework and test its predictions in financial markets. Whether the same correlation structure arises in other domains is an open empirical question.

\section{The Capability Paradox} 
\label{sec:paradox}

Throughout the paper, \emph{capability} refers to general-purpose reasoning and knowledge performance measured independently of our financial task. We use MMLU-Pro \cite{wang2024mmlu}, which evaluates factual and reasoning accuracy across domains, and ELO scores from LMSYS Chatbot Arena \citep{chiang2024chatbot}. Neither measure evaluates financial knowledge or trading performance. Capability scores and rankings are reported in Appendix~\ref{app:capability_benchmarks}.

In Section~\ref{sec:theory}, we introduced the risk floor $\bar\rho\,\sigma_r^2$. Capability can affect both terms. More capable models may extract richer shared features from their inputs, increasing $\bar\rho$, while improved reasoning may reduce residual variance $\sigma_r^2$. The risk floor rises when the increase in correlation dominates the reduction in residual variance. Standard individual benchmarks do not measure either quantity, so they can miss this system-level effect.

One possible explanation is that more capable models trained on overlapping data with similar objectives tend to identify the same latent features, so richer reasoning can homogenize behavior. This idea is similar to the Platonic Representation Hypothesis \citep{huh2024platonic}, though it has been recently challenged \citep{koepke2026cave}. Alternatively, RLHF and instruction tuning toward human-preferred outputs may push capable models toward a common response surface across architectures. In our work, we empirically show this convergence but leave future work to understand the mechanism behind it.

\section{Studying Systemic Risk}

In this section, we empirically validate the capability paradox presented in Section~\ref{sec:paradox}. We analyze individual model behavior and system-level scaling patterns across our three research questions.

\subsection{RQ1: Do frontier LLMs exhibit correlated behavior?}
\label{sec:rq1}

\begin{figure}[t]
    \centering
    \begin{subfigure}[t]{0.48\linewidth}
        \centering
        \includegraphics[width=\linewidth]{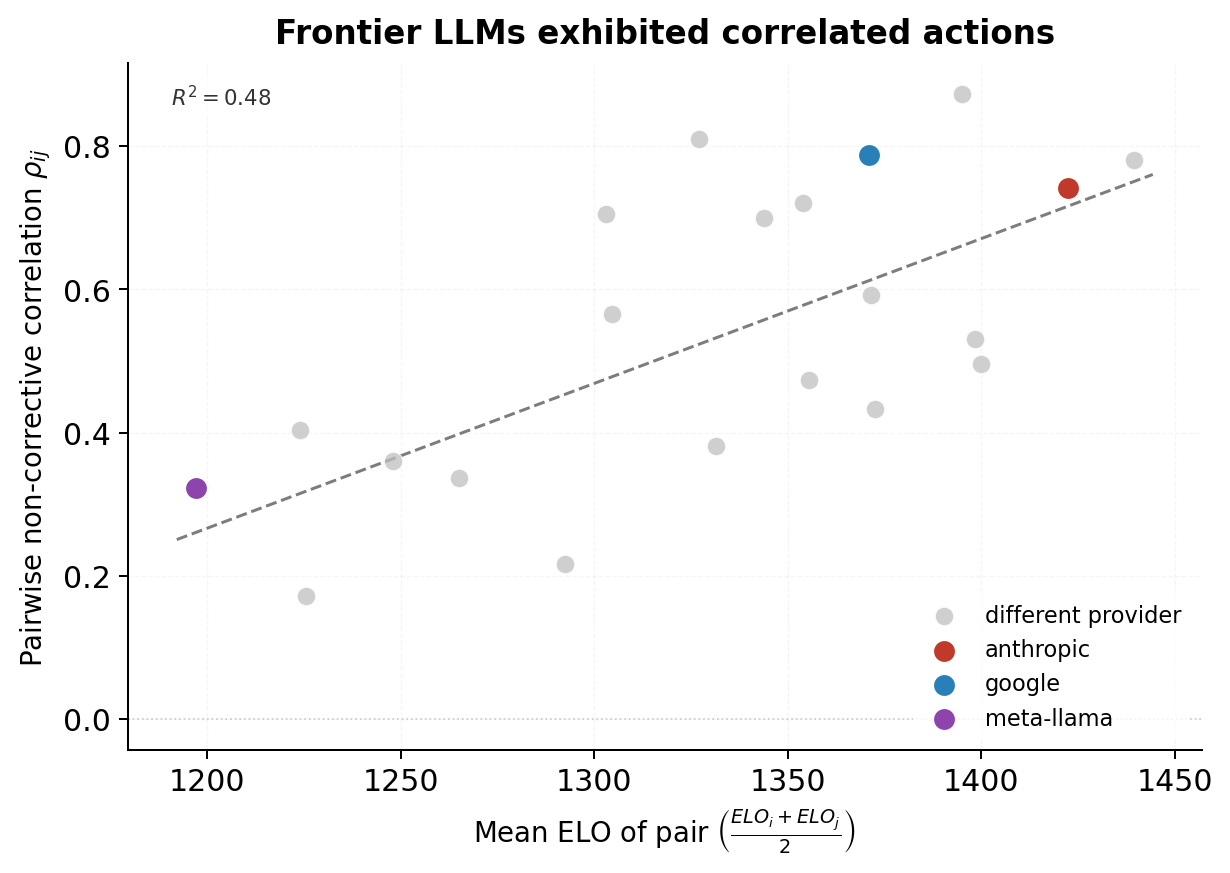}
        \caption{}
        \label{fig:rich_context}
    \end{subfigure}
    \hfill
    \begin{subfigure}[t]{0.48\linewidth}
        \centering
        \includegraphics[width=\linewidth]{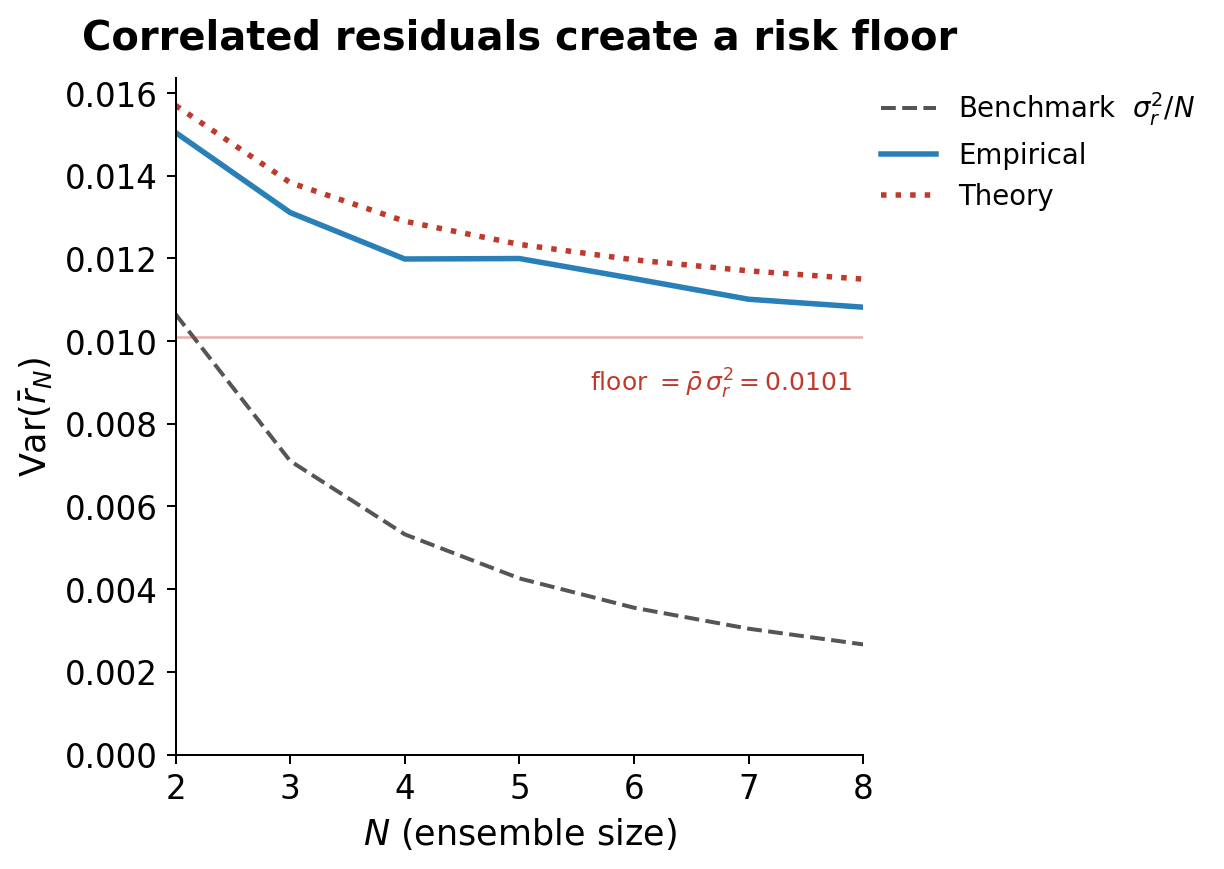}
        \caption{}
        \label{fig:empirics_theory}
    \end{subfigure}
    \caption{Mean pairwise non-corrective correlation vs.\ ELO capability score. Each point is a model; residuals are computed from a per-model OLS fit of actions on the fundamental gap, so correlation reflects the non-corrective component of behavior.}
    \label{fig:combined}
\end{figure}

To characterize what drives correlated behavior, we analyze model responses across systematically varied information conditions. To measure whether more capable models exhibit more correlated non-corrective trading behavior, we ran a controlled panel experiment in which 7 LLMs each responded to 64 scenarios with systematically varied fundamental gaps, momentum, volatility, and analyst views. For each model, we averaged actions across 5 samples per scenario, then fitted each model's action against the fundamental gap via OLS to isolate the component of behavior not explained by corrective price-gap responses. We then computed the pairwise Pearson correlation of the non-corrective components across all model pairs, yielding 21 pair-level observations. To test whether capability predicts this correlation, we regressed each pair's non-corrective correlation on the average capability score of the two models in the pair, controlling for whether they come from the same provider family. 

Capability scores significantly predict non-corrective correlated behavior between model pairs. In pairwise regressions estimating each pair's non-corrective correlation as a function of the pair's average capability score and whether the two models shared a provider, average capability was a strong, significant predictor under both ELO ($t = 3.94$, $p = 0.001$; $R^2 = 0.475$) and MMLU-Pro ($t = 4.88$, $p < 0.001$; $R^2 = 0.580$). Shared provider was not significant in either  specification ($p = 0.39$ for ELO, $p = 0.28$ for MMLU), which indicates that correlated behavior tracks raw capability more strongly than provider-specific tendencies.

We empirically validate the risk floor proposed in Section~\ref{sec:theory}. Even with a perfectly large ensemble, the variance of the ensemble-mean residual cannot fall below $\rho \cdot \sigma_r^2 = 0.010$, roughly half the single-model variance of 0.021. The empirical bootstrap (drawing N models with replacement, N = 2 to 8) closely tracks the theoretical curve, confirming that the floor is not an artifact of small samples. In contrast, an idealized benchmark assuming independent models would predict variance declining as $1/N$ with no floor at all. The practical implication is correlated errors among frontier LLMs create a persistent risk floor that diversification alone cannot eliminate.

These findings raise a natural next question: if capable LLMs exhibit correlated non-corrective behavior, what are the market-level consequences of deploying them at scale? We turn to this in RQ2.

\subsection{RQ2: Does increasing LLM market participation produce systemic risk?}

We now simulate a single-asset market in which three types of participants interact: LLM-based traders, noise traders, and a market maker. Prices are anchored to an exogenous fundamental value process --- that is, a true underlying asset value that evolves independently of agent actions --- which we control to create testable regime shifts (Figure~\ref{fig:simulation_fundamental}). At each timestep, all agents observe the current market state, and the reference price is updated from the resulting net order flow. Each condition is run with a homogeneous agent population with 5 independent seeds per condition.

\begin{figure}[t]
      \centering
      \begin{subfigure}[t]{0.48\linewidth}         
          \centering
          \includegraphics[width=\linewidth]{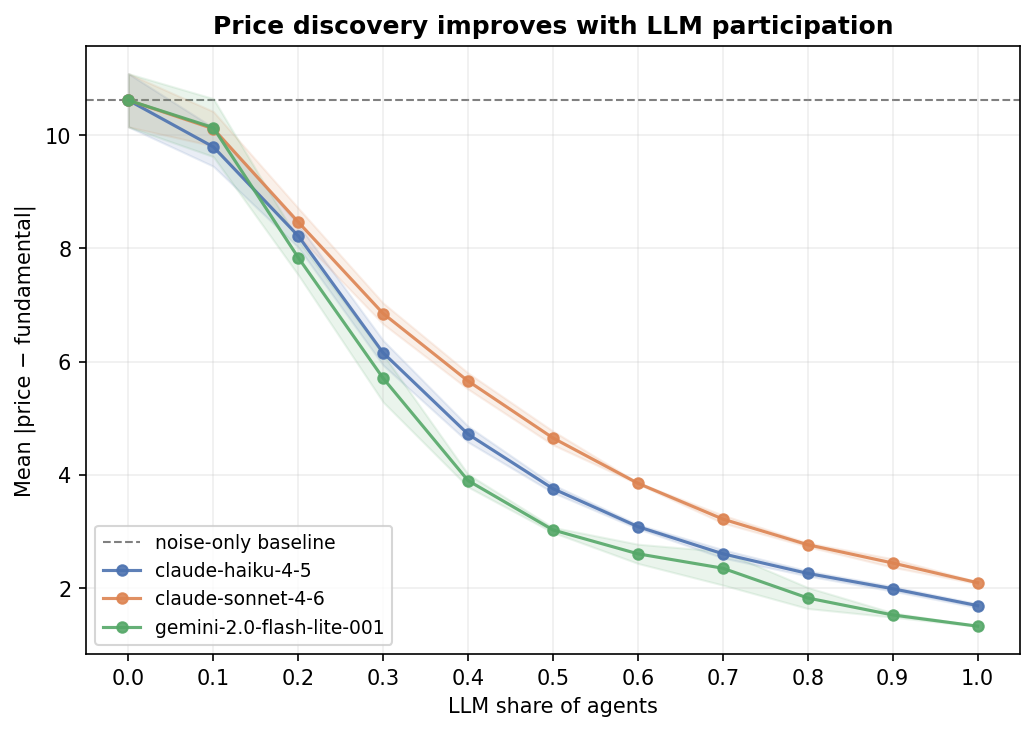}
          \caption{Mean tracking error (mean $|p_t - F_t|$) as a function of LLM participation share, averaged over seeds. All three models improve price discovery monotonically relative to the noise-only baseline (dashed). Shaded bands show $\pm 1$ SD across seeds.}
          \label{fig:paper_auc}                  
      \end{subfigure}
      \hfill
      \begin{subfigure}[t]{0.48\linewidth}
          \centering                             
          \includegraphics[width=\linewidth]{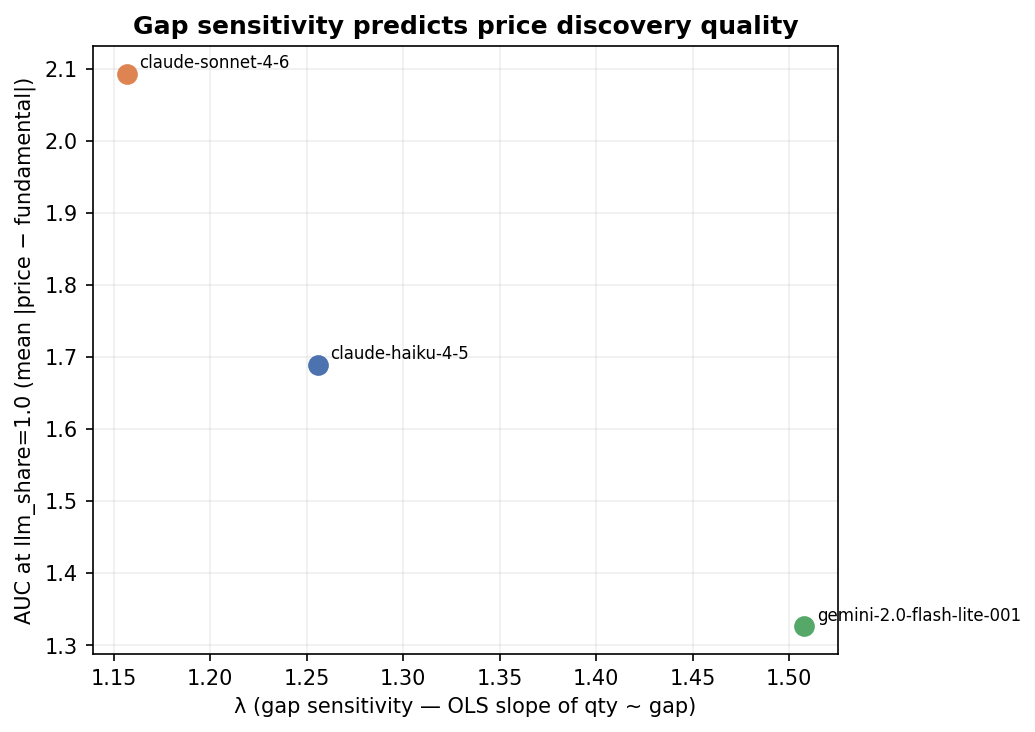}
        \caption{Gap sensitivity $\lambda$ versus steady-state tracking error at full LLM participation ($s{=}1.0$). Models with higher $\lambda$ achieve lower residual tracking error because they submit larger corrective orders per unit of mispricing.}
          \label{fig:paper_decomp}                
      \end{subfigure}
      \caption{LLM participation improves price discovery, and cross-model variation is explained by gap sensitivity. (Left) Tracking error falls monotonically as LLM share increases for all three model families. (Right) A single behavioral parameter---how  aggressively each model trades in response to perceived mispricing---accounts for the ranking of models at full participation.}
      \label{fig:rq2}                            
  \end{figure}   
  
\textbf{Setup.}
The fundamental value $F_t$ is exogenous and piecewise constant over 100 rounds, with two step shocks ($F: 100 \to 120$ at $t{=}33$; $F: 120 \to 90$ at $t{=}66$) that test whether prices track fundamentals through controlled discontinuities. In each simulation, the market is populated by one LLM family (Claude Haiku 4.5, Claude Sonnet 4.6, or Gemini 2.0 Flash Lite) alongside noise traders that draw signed quantities from a zero-mean distribution each period, providing a fundamental-free liquidity baseline. 

A market maker posts bid/ask quotes at a fixed 0.5\% spread around the reference price, which separates execution pricing from reference price evolution. After orders are submitted, the reference price updates as:
\[
P_t = P_{t-1} + \delta \cdot \frac{\mathrm{net}_t}{N},
\]
where $\mathrm{net}_t$ is total buy minus sell volume, $N$ is the number of agents, and $\delta$ is the price-impact parameter. Low $\delta$ represents a deep market (more liquid market in which prices are less sensitive to order flow); high $\delta$ a thin one (less liquid market with greater price responsivenes).

\textbf{Results.} Figure~\ref{fig:paper_auc} shows mean tracking error (mean $|p_t - F_t|$) as a function of LLM  share. The noise traders ($s=0$) provide a baseline mean tracking error. When LLMs replace these noise traders in the market, tracking error declines sharply and monotonically as LLM share increases. This confirms that LLM agents are effective arbitrageurs: they accurately perceive the fundamental gap and trade correctively, driving prices toward fair value more efficiently than noise traders alone.

Model families differ in how much price discovery they achieve at full participation. To explain this cross-model variation, we regress each agent's submitted quantity on the contemporaneous price--fundamental gap, estimating a gap sensitivity parameter $\lambda$ (OLS slope) for each model. Figure~\ref{fig:paper_decomp} reveals a near-perfect negative relationship between $\lambda$ and residual tracking error at $s{=}1.0$: Gemini 2.0 Flash Lite, which submits the largest corrective orders per unit of mispricing ($\lambda \approx 1.51$), achieves the lowest tracking error (${\approx}1.33$); Claude Sonnet 4.6, the most conservative trader ($\lambda \approx 1.15$), leaves the most residual error (${\approx}2.10$); and Claude Haiku 4.5 falls between the two ($\lambda \approx 1.25$, error ${\approx}1.69$). A single behavioral parameter---order aggressiveness in response to perceived mispricing---thus accounts for the ranking of model families along the price discovery dimension.

Taken together, these results suggest that correlated LLM behavior does not need to be destabilizing. When agents share accurate information, their agreement accelerates price discovery and limits distortion. What happens when that shared signal is wrong?

\subsection{RQ3: Does a shared information environment amplify systemic risk?}

Real financial markets increasingly expose LLM agents to common information streams---news feeds, analyst commentary, social media---that may temporarily diverge from fundamentals. To simulate this, we construct a shared information environment condition in which every agent, every round, receives eight pieces of market commentary unanimously asserting that the displayed fair-value signal overstates intrinsic value by 25--30\%. This condition simulates a scenario where agents are collectively exposed to a persistent, coherent narrative --- as might occur during a panic-driven news cycle or a viral bearish thesis --- that directly conflicts with the fundamental signal.

\textbf{Setup.} We extend the RQ2 simulation to a shared information environment by injecting distractor commentary into every agent's context at every round. Two conditions are compared: a neutral-sentiment baseline, in which eight pieces of rotating generic market commentary (e.g., "analysts are concerned," "the company beat earnings last quarter") are injected each round with mixed and inconsistent directional implications, and an adversarial condition in which eight pieces of rotating commentary unanimously assert that the displayed fair-value signal overstates intrinsic value by 25--30\%, instructing agents to trade on the adjusted, lower estimate. Adversarial distractors directly impugn the reliability of the signal itself, providing agents with a coherent and internally consistent rationale for discounting it.

Agent composition and market mechanics are otherwise identical to RQ2. In this experiment, Claude Haiku 4.5, Claude Sonnet 4.6 and Gemini 2.0 Flash are independently mixed with noise traders at participation shares $s \in \{0.0, 0.2, 0.4, 0.6, 0.8, 1.0\}$ under the full market context regime, with 3 independent seeds per condition. The fundamental value schedule follows the same two-shock structure ($v\colon 100 \to 120$ at $t{=}33$; $v\colon 120 \to 90$ at $t{=}66$). Because every agent draws from the same rotating distractor pool each round, any reasoning errors induced by the commentary are fully correlated across agents --- there is no diversification of the misinformation. This is a direct consequence of the population-shared component $z^{\text{port}}_t$ in the residual decomposition (Appendix~\ref{app:noise_decomposition}): when all agents draw from the same information environment, the shared error term dominates and population scaling provides no relief.

\begin{figure}[t]
    \centering
    \includegraphics[width=\linewidth]{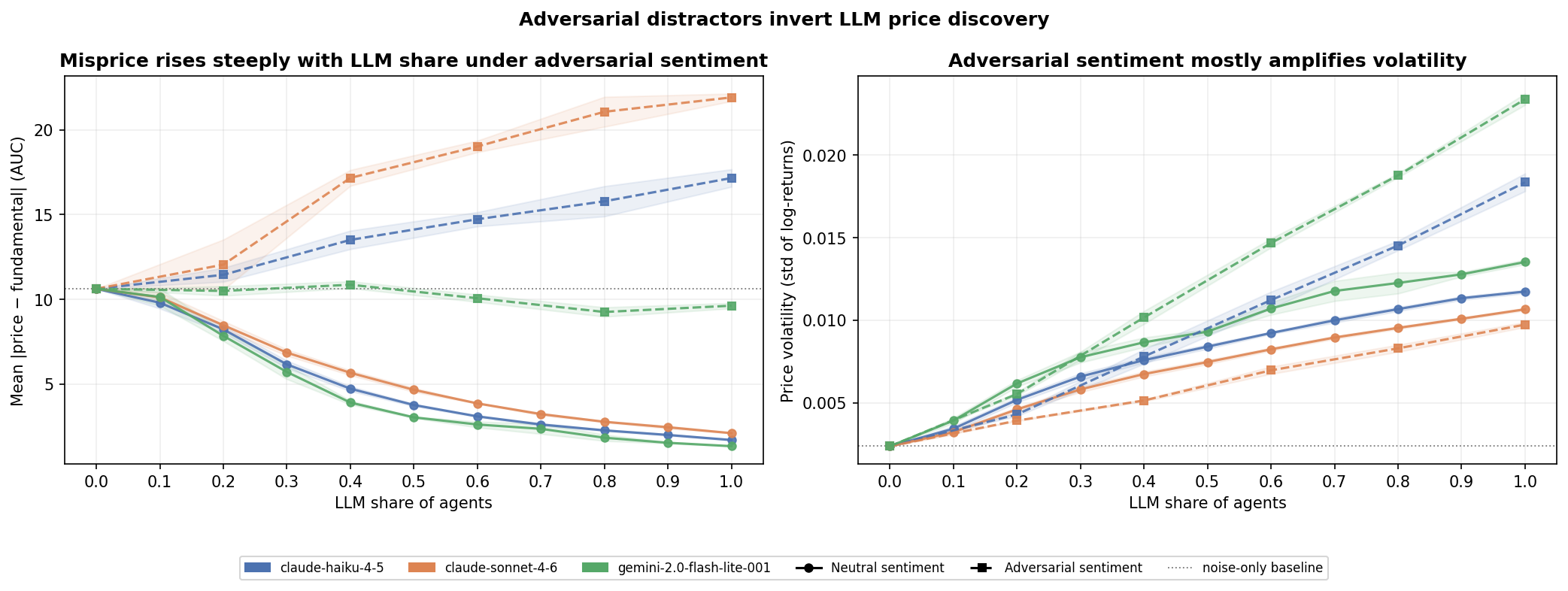}
    \caption{Mean tracking error (left) and realized volatility (right) as a function of LLM participation share under two distractor regimes. Adversarial distractors, which directly impugn the fundamental signal, invert the corrective benefit of LLM participation seen in RQ2. Shaded bands show $\pm 1$ SD across seeds.}
    \label{fig:rq3}
\end{figure}

\textbf{Results.} The results sharply qualify the 
conclusion from RQ2. Figure \ref{fig:rq3} shows that adversarial distractors dramatically degrade price discovery, and that the damage scales with LLM market share. Under neutral sentiment, all three models track the fundamental value better than the noise-only baseline regardless of LLM market share, which indicates that generic commentary does not materially impair inference. Under adversarial sentiment, however, mean tracking error rises steeply with LLM share, reaching five to seven times the baseline at full LLM penetration --- a near-inversion of the price discovery function. Volatility follows a similar pattern, increasing with LLM share in both conditions but more sharply under adversarial distractors for Haiku and Gemini; Sonnet shows a smaller volatility gap between conditions. Together, these results suggest that LLM agents are susceptible to coordinated misinformation attacks on their valuation anchor: when the fundamental signal is discredited, LLMs actively trade against fundamentals, transforming market stabilizers into destabilizers.

The mechanism is a correlated signal inversion. When the true fundamental jumps, noise traders leave the market underpriced, creating a strong buy signal. Under adversarial distractors, agents follow the commentary's instruction to treat the fair-value signal as overstated, infer that the stock is overpriced, and sell. At full LLM penetration, directional accuracy collapses to 54\% for Haiku and 68\% for Gemini; Sonnet falls to 42\%, below chance, meaning its orders are net contrarian. Under neutral distractors, all three models follow the gap and achieve 100\% directional accuracy --- every non-zero order moves price toward fair value. Crucially, because all agents draw from the same information environment, these errors are fully correlated: each additional LLM agent amplifies the distortion.

This result isolates a distinct failure mode from RQ2. There, LLM participation improved average price discovery while amplifying peak instability --- a mixed outcome. Here, a shared misinformation environment eliminates the stabilizing benefit entirely, producing a market that is strictly worse than one populated by uninformed noise traders alone. The implication is that correlated reasoning errors, induced by common context, pose a qualitatively different and more severe systemic risk than correlated noise.

\section{Related Work}
\label{sec:related_work}

The literature on LLM capability scaling characterizes individual model performance but does not ask how system-level outcomes change when capable agents operate collectively in dynamic environments. The algorithmic monoculture literature establishes that correlated errors are a structural consequence of shared models, but studies them in static, one-shot settings. The financial systemic risk literature shows how individually rational behavior generates collective fragility, but does not consider LLM-specific failure modes. We contribute a framework that connects all three: we show analytically that in dynamic feedback systems, correlated LLM errors are non-diversifiable as the population grows and that this non-diversifiable noise interacts with individual capability in a non-monotone way.

\textbf{Capability Scaling.} Kaplan et al. \cite{kaplan2020scaling} established that LLM loss follows power-law relationships with compute, data, and parameters, while Pearce et al. \cite{pearce2024scalinglawspretrainingagents} refined these to compute-optimal regimes. These laws characterize capability in isolation, a single model evaluated on static benchmarks. A growing literature documents the points where individual capability scaling breaks down: McKenzie et al. \cite{mckenzie2023inverse} identified tasks where performance degrades monotonically with scale, and Wei et al. \cite{wei2023inverse} showed that many such tasks exhibit U-shaped scaling at sufficient scale: performance first worsens then recovers. Schaeffer et al. \cite{schaeffer2023emergent} argued that nonlinear metrics can create apparent phase transitions in capability. Ruan et al. \cite{ruan2024observational} demonstrated that complex agentic behaviors can nonetheless be predicted from simpler capability measures via observational scaling laws. Our work extends this program to system-level scaling: we study how outcomes change when capable agents operate collectively in dynamic environments.


\textbf{Algorithmic Monoculture.} Kleinberg and Raghavan~\citep{kleinberg2021algorithmic} proved that shared adoption of a single algorithm can reduce social welfare even when that algorithm is individually superior, due to correlated errors across decision-makers. Bommasani et al.~\citep{bommasani2021opportunities} identified homogenization as a structural risk of foundation models, and subsequent work by Bommasani et al.~\citep{bommasani2022picking} and Toups et al.~\citep{toups2023ecosystem} demonstrated empirically that shared model families produce homogeneous outcomes at ecosystem scale. Kim et al.~\citep{kim2025correlated} measured error agreement across 350+ LLMs and found roughly 60\% agreement on incorrect answers, driven by shared architectures; Goel et al.~\citep{goel2025great} showed that this agreement increases with capability, undermining AI oversight. Huh et al.~\citep{huh2024platonic} provided a mechanistic account: as models scale, their representations converge toward a shared statistical model of reality, making correlated failure modes an intrinsic consequence of scaling. This literature characterizes correlated errors in \textit{static} settings, that is, in individual decisions, fixed benchmarks, and one-shot evaluations. Our setting adds dynamic feedback: correlated agents update a shared state, allowing errors to compound over time.

\textbf{Financial Systemic Risk.}
Academic finance has long studied how individually rational actions can generate system-level fragility. A standard taxonomy distinguishes three channels: contagion, in which initially localized distress propagates through financial linkages; common-exposure risk, in which many institutions are simultaneously affected by the same shock; and endogenous imbalance, in which leverage, liquidity dependence, or crowded positions build up over time and unwind abruptly. Canonical models study contagion through interbank exposures and clearing networks \citep{allen_financial_2000,eisenberg_systemic_2001}, amplification through financial networks \citep{acemoglu2015systemic,elliott_financial_2014}, and correlated institutional positions that make individually diversified actors jointly vulnerable to common shocks \citep{wagner_diversification_2010}.

Our paper is closest to the common-exposure channel. We study a behavioral aggregation mechanism: as institutions adopt similar models for information processing, trading, risk management, and decision support, model-driven behavior may become more homogeneous, harder to supervise, and more reliant on unreliable outputs \citep{cecchetti_artificial_2025}. By showing how shared LLM behavior can create a non-diversifiable residual component in aggregate action, we connect systemic-risk theory to LLM evaluation: investigating individual agent capabilities and whether populations of capable agents remain dynamically stable at scale.

\section{Discussion}
\label{sec:discussion}
Standard model evaluation asks whether a single deployment improves on a baseline. Systemic safety also depends on how deployed models behave together. A model can achieve state-of-the-art performance on every standard evaluation while contributing a large aggregation floor to every system it is deployed in, and no current audit would detect it. The most uncomfortable implication of our results is that the natural  response to correlated failures --- deploying more diverse models from more providers --- does not address the dominant driver. Sharing a provider family does not predict non-corrective correlation: two high-capability models from different companies are more correlated with each other than two low-capability models from the same company. 

This matters most in the tail. Under adversarial information environments, LLM participation produces tracking error five to seven times the noise-only baseline --- because their failures are synchronized. The aggregation framework predicts that correlated residual risk persists as the population grows. Our experiments establish the corresponding failure mode in simulated financial markets; population-size scaling remains an important empirical test.

The risk of correlated algorithmic behavior predates LLMs: the 2010 Flash Crash showed how algorithms sharing similar heuristics can synchronize catastrophically, briefly erasing nearly \$1 trillion in market value \citep{kirilenko2017flash}. However, the risk of LLM monoculture is harder to observe because it arises from shared implicit priors through shared implicit priors. Just as financial regulators monitor systemic exposure across institutions alongside institution-level evaluation, AI governance needs population-level correlation audits alongside individual model evaluations. An open question is how to decorrelate model populations without sacrificing individual capability --- whether through training for behavioral diversity, input perturbation, or regulatory requirements for model provenance disclosure. There is also a collective action problem: individual deployers are incentivized to use the most capable available model, even as aggregate welfare would be better served by a more heterogeneous population. Resolving this tension may require coordination mechanisms that go beyond what any single deployer can achieve unilaterally.

\textbf{Limitations.} All empirical evidence in this paper comes from simulated financial markets. The setting gives us a measurable fundamental value and controlled information shocks, but it remains a simplified single-asset market with exogenous fundamentals and stylized liquidity. Real markets contain correlated assets, endogenous liquidity, strategic adaptation, and institutional constraints that could amplify or dampen the effects we document. Whether the same correlation structure and feedback dynamics arise in other domains is an open empirical question.

\section{Conclusion}

We set out to ask whether improving individual model capability can degrade rather than improve system-level outcomes. In our financial market simulation, it can. More capable models exhibit more correlated non-corrective behavior, accurate shared reasoning improves price discovery, and a shared misinformation environment turns the same coordination into a liability. The resulting capability paradox is simple: individual performance improvements do not necessarily translate into better market-level outcomes when model behavior is correlated.

These results motivate population-level evaluation alongside individual model evaluation. We present a general framework and test its predictions in financial markets. Whether the same dynamics arise in other domains is an open empirical question.

\bibliographystyle{plainnat}
\bibliography{ref}

\newpage
\appendix
\section{Application to Insurance Billing in Medical Settings}
\label{app:insurance_billing}

The healthcare billing ecosystem provides a natural adversarial analogue to our framework. Provider-side LLM agents (e.g., AI scribes) optimize documentation and coding to maximize reimbursement, while payer-side LLM agents (insurance auditing systems) optimize claim review to minimize payouts. Both act on a shared system state --- aggregate billing intensity --- relative to an underlying ground truth of appropriate care and billing.

\paragraph{System mapping.}
The system state $S_t$ represents aggregate billing intensity at time $t$: the prevailing level of reimbursement claims relative to underlying care delivered. Ground truth $F$ is the appropriate billing level --- what would be claimed under accurate coding and valid denial decisions. Provider-side agents take actions that push $S_t$ upward (toward higher reimbursement); payer-side agents take actions that push $S_t$ downward (toward lower payouts). The state evolves as in Equation~\ref{eq:state_evolution}, with the gap $G_{t-1} = F - S_{t-1}$ measuring the deviation of prevailing billing intensity from ground truth.

\paragraph{Corrective and non-corrective components.}
Applying the decomposition of Equation~\ref{eq:decomp}, each agent's action splits into a corrective component $\lambda_i G_{t-1}$ and a non-corrective residual $r_t^{(i)}$. For a provider-side agent, the corrective component captures responses to genuine undercoding --- legitimate gaps between care delivered and reimbursement claimed. For a payer-side agent, it captures responses to genuine overbilling --- valid grounds for claim denial. In both cases the non-corrective residual captures everything else: responses to exploitable ambiguities in billing rules, surface patterns in documentation that do not track appropriate care, or heuristics inherited 
from training data that activate regardless of whether a gap exists.

\paragraph{Correlated non-corrective action and the aggregation floor.}
Because provider-side and payer-side models are often trained on similar corpora and share architectural biases, their non-corrective residuals $r_t^{(i)}$ can become correlated within each side. A shared heuristic --- for instance, a learned association between particular diagnostic codes and reimbursement outcomes --- activates the same non-corrective response across many provider agents simultaneously. 
By the result in Section~\ref{sec:theory}, this positive $\bar\rho$ creates an aggregation floor $\bar\rho\,\sigma_r^2$ that cannot be eliminated by deploying more agents. Correlated non-corrective actions compound rather than cancel in aggregate.

\paragraph{Endogenous adversarial feedback.}
Unlike the symmetric settings of financial markets or medical diagnosis, this setting introduces a complication absent from the main framework: \emph{endogenous adversarial feedback}. One side's non-corrective behavior shapes the distribution of inputs the other side learns to respond to. If provider agents learn to exploit ambiguities in billing rules, payer agents trained on the resulting claims distribution learn to deny more aggressively, shifting their own non-corrective residuals in response. This 
feedback loop means that capability improvements do not straightforwardly reduce non-corrective behavior: a more capable provider agent extracts richer features from billing rules, improving its corrective component on clear-cut cases while also becoming more effective at identifying and acting on exploitable ambiguities, enlarging $r_t^{(i)}$ endogenously. The capability paradox of Section~\ref{sec:theory} is therefore strengthened in this setting: individual capability improvements can raise $\bar\rho$ on both sides simultaneously, amplifying the aggregation floor even as each individual agent becomes more accurate on its own benchmark.

\paragraph{The dangerous regime.}
The highest-risk scenario combines large non-corrective residuals on both sides with high within-side correlation and adversarial coupling across sides. In this regime, correlated provider agents inflate billing intensity together while correlated payer agents deny claims together, producing systemic instability --- billing inflation, 
denial inflation, and increased administrative overhead --- that individual evaluation of either side cannot detect. As in the main framework, neither individual accuracy nor population size alone identifies the risk: the relevant objects are $\bar\rho$ and $\sigma_r^2$ on each side, and how adversarial feedback causes them to co-evolve.

\section{Residual Decomposition and Sources of Correlated Action}
\label{app:noise_decomposition}

We decompose the non-corrective residual into three components to make the sources of correlated behavior explicit:
\begin{equation}
    r_t^{(i)} = z^{\text{arch}}_{t,\,f(i)} + z^{\text{port}}_t + 
    \xi_t^{(i)},
    \label{eq:residual_decomp}
\end{equation}
where $f(i)$ is the model family of agent $i$, $z^{\text{arch}}_{t,\,f(i)}$ 
is a mean-zero family-shared component capturing shared priors from 
architecture and training (variance $\sigma_a^2$), $z^{\text{port}}_t$ is 
a mean-zero population-shared component capturing shared environmental 
state that affects all agents regardless of family (variance $\sigma_p^2$), 
and $\xi_t^{(i)}$ is agent-specific residual noise (variance $\sigma_\xi^2$). 
The three components are mutually uncorrelated and family-shared components 
are independent across families. Total residual variance is 
$\sigma_r^2 = \sigma_a^2 + \sigma_p^2 + \sigma_\xi^2$.

\paragraph{Single-family case.}
For a population in which all agents are from the same family $f$, the average pairwise residual correlation is
\begin{equation}
    \bar\rho_w = \frac{\sigma_a^2 + \sigma_p^2}{\sigma_a^2 + \sigma_p^2 + 
    \sigma_\xi^2}.
\end{equation}

\paragraph{Multi-family case.}
For a mixed population, residual correlation is family-pair-specific:
\begin{equation}
    \mathrm{Corr}(r^{(i)}_t, r^{(j)}_t) =
    \begin{cases}
        \bar\rho_w & \text{if } f(i) = f(j),\; i \neq j, \\[4pt]
        \rho_c := \dfrac{\sigma_p^2}{\sigma_a^2 + \sigma_p^2 + \sigma_\xi^2} 
        & \text{if } f(i) \neq f(j).
    \end{cases}
\end{equation}
The cross-family correlation $\rho_c$ is exactly the share of total residual variance attributable to the population-shared component. Family 
mixing eliminates $\sigma_a^2$ from cross-family pairs but leaves 
$\sigma_p^2$ untouched. In our market simulation, the dominant source of 
$z^{\text{port}}_t$ is symmetric portfolio initialization: agents reach 
inventory or cash limits at similar moments and respond similarly 
regardless of model family. This means that even a fully 
provider-diversified deployment retains a non-diversifiable floor 
$\bar\rho_c\,\sigma_r^2 > 0$ so long as $\sigma_p^2 > 0$.

\paragraph{Implication for RQ3.}
The shared information environment in RQ3 is a large positive shock to 
$z^{\text{port}}_t$: the distractor commentary is drawn from the same 
pool by every agent every round, so any reasoning errors it induces are 
fully correlated across agents regardless of model family. This is why 
the adversarial condition produces tracking error that scales with LLM 
share rather than averaging out.

\section{Simulation Details and Scope}
\label{app:additional_modeling}

\paragraph{Simulation details.}
The fundamental value is exogenous and piecewise constant with regime 
shifts at prespecified times. LLM agents observe current price, 
short-horizon price history, their cash and inventory state, and a noisy 
fair-value signal; they output buy/sell/hold decisions and integer 
quantities. Noise traders submit signed quantities drawn from a zero-mean 
process, providing an independent baseline with $\bar\rho = 0$ by 
construction. The market maker quotes a fixed bid/ask spread around the 
reference price, separating execution pricing from reference price 
evolution. The reference price updates from normalized net order imbalance:
\begin{equation}
    P_t = P_{t-1} + \delta \cdot \frac{\mathrm{net}_t}{N},
\end{equation}
where $\delta$ controls effective market depth: low $\delta$ corresponds 
to deeper markets where prices are less sensitive to order flow; high 
$\delta$ to thinner markets where the same flow moves prices more.

\paragraph{Scope conditions.}
Three conditions bound the applicability of the AR(1) state-variance 
predictions. First, regime shifts in the fundamental value create tracking 
transients on top of the stationary residual-variance component; the floor 
characterizes the noise component, not the transient. Second, positive 
serial correlation in residuals increases long-run variance relative to 
the i.i.d.\ benchmark, so the floor is a lower bound in this case. Third, 
cash and inventory constraints make the linear state-variance predictions 
approximate when constraints bind frequently; the AR(1) derivation assumes 
unconstrained agents.

\section{Capability Benchmarks}
\label{app:capability_benchmarks}

We report the MMLU scores \cite{wang2024mmlu} for models used in experiments RQ1-3 in Figure \ref{tab:capability_benchmarks}.

\begin{table}[h]
\centering
\caption{Capability benchmarks for the 8 models used in our panel experiments. MMLU-Pro scores are from Open LM Arena \cite{chiang2024chatbot}; ELO \cite{chiang2024chatbot}.}
\label{tab:capability_benchmarks}
\begin{tabular}{llrr}
\toprule
\textbf{Model} & \textbf{Organization} & \textbf{MMLU-Pro (\%)} & \textbf{ELO} \\
\midrule
Llama-3.2-3B-Instruct          & Meta      & 34.7 & 1118 \\
gpt-5.4-nano              & OpenAI    & 35.6 & 1406 \\
Gemini-2.0-Flash-Lite          & Google    & 72.4 & 1330 \\
Llama-4-Scout-17B-16E-Instruct & Meta      & 75.2 & 1276 \\
GPT-5-nano                     & OpenAI    & 77.2 & 1333 \\
Claude Haiku 4.5               & Anthropic & 80.0 &  1378 \\
Gemini-2.5-Flash               & Google    & 83.2 & 1412 \\
Claude Sonnet 4.6              & Anthropic & 88.0 & 1467 \\
\bottomrule
\end{tabular}
\end{table}

\section{LLM Usage}

Claude Code was used to assist the authors in implementing the market simulation and analysis. Claude was also used to polish the writing of the paper. 

\section{External Assets and Licenses}

The experiments use commercial and open model APIs accessed through authorized accounts, public benchmark scores for capability measurement, and author-written simulation and analysis code. We do not redistribute model weights or proprietary provider assets. Model outputs used in the experiments were generated by the authors from the prompts described in the paper and supplemental material. Public benchmark sources, including MMLU-Pro and Chatbot Arena/Open LM Arena scores, are cited where used. All released code and experimental artifacts will be distributed with a repository license and will retain attribution for any third-party software dependencies.

\end{document}